\pdfoutput=1

\documentclass[11pt]{article}

\usepackage{EMNLP2022}

\usepackage{times}
\usepackage{latexsym}

\usepackage[T1]{fontenc}

\usepackage[utf8]{inputenc}

\usepackage{microtype}

\usepackage{inconsolata}

\usepackage{amsmath}
\DeclareMathOperator*{\argmin}{arg\,min}
\usepackage{booktabs}
\usepackage{graphicx}
\usepackage{xcolor}
\usepackage{placeins}
\usepackage{float}
\usepackage{caption}
\usepackage{subcaption}
\usepackage{cleveref}
\usepackage{tabularx}
\title{Are Representations Built from the Ground Up? \\
An Empirical Examination of Local Composition in Language Models}

\author{Emmy Liu \and Graham Neubig \\
Language Technologies Institute \\
Carnegie Mellon University \\
\texttt{\{mengyan3, gneubig\}@cs.cmu.edu}}

\begin{document}
\maketitle
\begin{abstract}
\emph{Compositionality}, the phenomenon where the meaning of a phrase can be derived from its constituent parts, is a hallmark of human language. At the same time, many phrases are \emph{non-compositional}, carrying a meaning beyond that of each part in isolation. Representing both of these types of phrases is critical for language understanding, but it is an open question whether modern language models (LMs) learn to do so; in this work we examine this question. We first formulate a problem of predicting the LM-internal representations of longer phrases given those of their constituents. We find that the representation of a parent phrase can be predicted with some accuracy given an affine transformation of its children. While we would expect the predictive accuracy to correlate with human judgments of semantic compositionality, we find this is largely \emph{not} the case, indicating that LMs may not accurately distinguish between compositional and non-compositional phrases. We perform a variety of analyses, shedding light on when different varieties of LMs do and do not generate compositional representations, and discuss implications for future modeling work.\footnote{Code and data available at \url{https://github.com/nightingal3/lm-compositionality}}
\end{abstract}

\section{Introduction}

Compositionality is argued to be a hallmark of linguistic generalization \cite{sep-compositionality}.
However, some phrases are non-compositional, and cannot be reconstructed from individual constituents \cite{dankers2021-paradox}.  Intuitively, a phrase like "I own cats and dogs" is locally compositional, whereas "It's raining cats and dogs" is not. Therefore, any representation of language must be easily composable, but it must also correctly handle cases that deviate from compositional rules.

\begin{figure}[!ht]
    \centering
    \includegraphics[width=0.65\columnwidth]{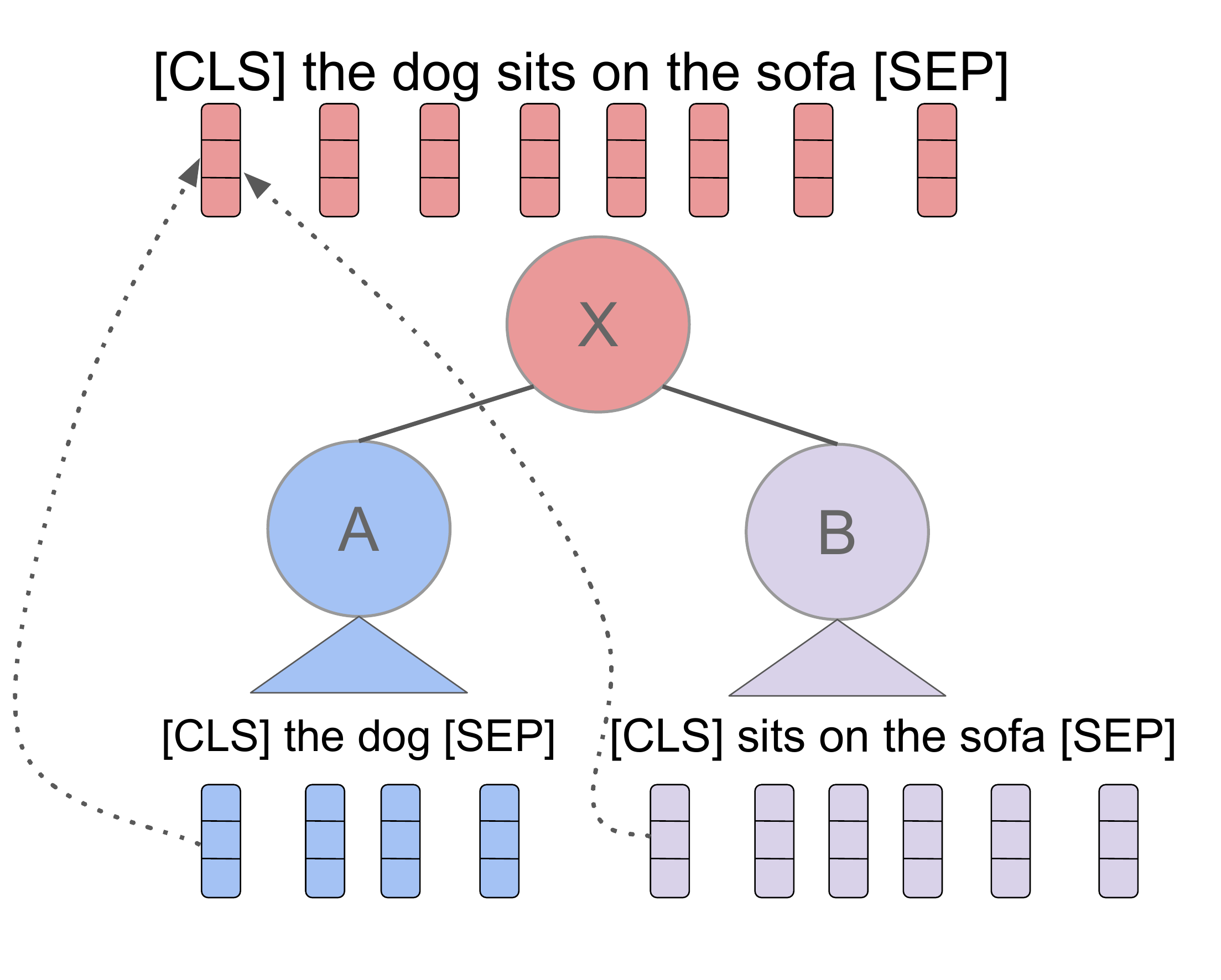}
    \caption{An illustration of the local composition prediction problem with \texttt{[CLS]} representations.}
    \label{fig:illustr}
\end{figure}

Both lack \cite{hupkes-review, Lake2017StillNS} and excess \cite{dankers-etal-2022-transformer} of compositionality have been cited as common sources of errors in NLP models, indicating that models may handle phrase composition in an unexpected way.

In general form, the compositionality principle is simply ``the meaning of an expression is a function of the meanings of its parts and of the way they are syntactically combined'' \cite{pelletier1994}. However, this definition is underspecified \cite{partee1984}. Recent efforts to evaluate the compositional abilities of neural networks have resulted in several testable definitions of compositionality \cite{hupkes-review}. 

Previous work on compositionality in natural language focuses largely on the definition of \textbf{substitutivity}, by focusing on changes to the constituents of a complex phrase and how they change its representation \cite{dankers2021-paradox, garcia-etal-2021-probing, Yu2020AssessingPR}. The definition we examine is \textbf{localism}: whether or not the representation of a complex phrase is derivable only from its local structure and the representations of its immediate ``children'' \cite{hupkes-review}. A similar concept has been proposed separately to measure the compositionality of learned representations, which we use in this work \cite{andreas2019measuring}.  We focus on localism because it is a more direct definition and does not rely on the collection of contrastive pairs of phrases. This allows us to examine a wider range of phrases of different types and lengths. 

In this paper, we ask whether reasonable compositional probes can predict an LM's representation of a phrase from its children in a syntax tree, and if so, which kinds of phrase are more or less compositional. We also ask whether this corresponds to human judgements of compositionality. 

We first establish a method to examine local compositionality on phrases through probes that try to predict the representation of a parent given its children (\cref{sec:methods}). We create two English-language datasets upon which to experiment: a large-scale dataset of 823K phrases mined from the Penn Treebank, and a new dataset of idioms and paired non-idiomatic phrases for which we elicit human compositionality judgements, which we call the \textbf{C}ompositionality of \textbf{H}uman-annotated \textbf{I}diomatic \textbf{P}hrases dataset (\textbf{CHIP}) (\cref{sec:data}). 

For multiple models and phrase types, we find that phrase embeddings across models and representation types have a fairly predictable affine compositional structure based on embeddings of their constituents (\cref{sec:comp-fn}). We find that there are significant differences in compositionality across phrase types, and analyze these trends in detail, contributing to understanding how LMs represent phrases (\cref{sec:phrase-types}). Interestingly, we find that human judgments do not generally align well with the compositionality level of model representations (\cref{sec:human-results}). This implies there is still work to be done at the language modelling level to capture a proper level of compositionality in representations.

\section{Methods and Experimental Details}
\label{sec:methods}
\subsection{Tree Reconstruction Error}
We follow \citet{andreas2019measuring} in defining deviance from compositionality as \textit{tree reconstruction error}.
Consider a phrase $x = [a][b]$, where $a$ and $b$ can be any length $> 0$. Assume we always have some way of knowing how $x$ should be divided into $a$ and $b$. Assume we also have some way of producing representations for $x$, $a$, and $b$, which we represent as a function $r$. Given representations $r(x)$, $r(a)$ and $r(b)$, we wish to find the function which most closely approximates how $r(x)$ is constructed from $r(a)$ and $r(b)$.

\begin{align}
    \hat{f} = \argmin_{f \in \mathcal{F}} \frac{1}{|\mathcal{X}|}\sum_{x \in \mathcal{X}} \delta_{x, ab} \\
    \delta_{x, ab} = d(r(x), f(r(a), r(b))
\end{align}

Where $\mathcal{X}$ is the set of possible phrases in the language that can be decomposed into two parts, $\mathcal{F}$ is the set of functions under consideration, and $d$ is a distance function. An example scenario is depicted in \autoref{fig:illustr}.

For $d$, we use cosine distance as this is the most common function used to compare semantic vectors. The division of $x$ into $a$ and $b$ is specified by syntactic structure \cite{chomsky-normal-form}. Namely, we use a phrase's annotated constituency structure and convert its constituency tree to a binary tree with the right-factored Chomsky Normal Form conversion included in NLTK \cite{nltk}.

\subsection{Language Models}

We study representations produced by a variety of widely used language models, specifically the \texttt{base-(uncased)} variants of Transformer-based models: \textbf{BERT}, \textbf{RoBERTa}, \textbf{DeBERTa}, and \textbf{GPT-2} \cite{deberta, roberta, bert, gpt2}.

\subsubsection{Representation extraction}
Let $[x_0, ..., x_{N}]$ be a sequence of $N + 1$ input tokens, where $x_0$ is the \texttt{[CLS]} token if applicable, and $x_{N}$ is the end token if applicable. Let $[h^{(i)}_0, ..., h^{(i)}_{N}]$ be the embeddings of the input tokens after the $i$-th layer. 

For models with the \texttt{[CLS]} beginning of sequence token (BERT, RoBERTa, and DeBERTa), we extracted the embedding of the \texttt{[CLS]} token from the last layer, which we refer to as the \textbf{CLS} representation. For GPT-2, we extracted the last token, which serves a similar purpose. This corresponds to $h^{(12)}_{0}$ and $h^{(12)}_{N}$ respectively.

Alternately, we also averaged all embeddings from the last layer, including special tokens. We refer to this as the \textbf{AVG} representation.
 
\begin{align}
    \frac{1}{N+1} \sum_{i = 0}^{N + 1} h^{(12)}_i
\end{align}

\subsection{Approximating a Composition Function}

To use this definition, we need a composition function $\hat{f}$. We examine choices detailed in this section.

For parameterized probes, we follow the probing literature in training several probes to predict a property of the phrase given a representation of the phrase. However, in this case, we are not predicting a categorical attribute such as part of speech. Instead, the probes that we use aim to predict the parent representation $r(x)$ based on the child representations $r(a)$ and $r(b)$. We call this an \textit{approximative probe} to distinguish it from the usual use of the word probe.

\subsubsection{Arithmetic Probes}

In the simplest probes, the phrase representation $r(x)$ is computed by a single arithmetic operation on $r(a)$ and $r(b)$. We consider three arithmetic probes:\footnote{Initially, we considered the elementwise product \textbf{PROD($r(a)$, $r(b)$)} $= r(a) \odot r(b)$, but found that it was an extremely poor approximation.}
\begin{align}
    \textbf{ADD}(r(a), r(b)) = r(a) + r(b) \\
    \textbf{W1}(r(a), r(b)) = r(a) \\ 
    \textbf{W2}(r(a), r(b) = r(b)
\end{align}

\subsubsection{Learned Probes}

We consider three types of learned probes. The linear probe expresses $r(x)$ as a linear combination of $r(a)$ and $r(b)$. The affine probe adds a bias term. The MLP probe is a simple feedforward neural network with 3 layers, using the ReLU activation.
\begin{align}
    \textbf{LIN}(r(a), r(b)) = \alpha_1 r(a) + \alpha_2 r(b) \\
    \textbf{AFF($r(a)$, $r(b)$)} = \alpha_1 r(a) + \alpha_2 r(b) + \beta \\
    \textbf{MLP}(r(a), r(b)) = W_3h_2
\end{align}

Where 
$$h_1 = \sigma(W_1 [r(a);r(b)])$$
$$h_2 = \sigma(W_2h_1),$$
$W_1$ is $(300 \times 2)$, $W_2$ is $(768 \times 300)$, and $W_3$ is $(1 \times 768)$. We do not claim that this is the best MLP possible, but use it as a simple architecture to contrast with the linear models. 


\section{Data and Compositionality Judgments}
\label{sec:data}

\subsection{Treebank}
To collect a large set of phrases with syntactic structure annotations, we collected all unique subphrases ($\geq 2$ words) from WSJ and Brown sections of the Penn Treebank (v3) \cite{treebank}. \footnote{We converted the trees to Chomsky Normal Form with right-branching using NLTK \cite{nltk}. We note that not all subtrees are syntactically meaningful. However, we used this conversion to standardize the number of children and formatting. We exclude phrases with a null value for the left or right branch \cite{treebank-guidelines}.}

The final dataset consists of \textbf{823K} phrases after excluding null values and duplicates.  We collected the length of the left child in words, the length of the right child in words, and the tree's production rule, which we refer to as \textit{tree type}. There were 50260 tree types in total, but many of these are unique. Examples and phrase length distribution can be found in \autoref{sec:treebank-tree-types}, and \autoref{sec:treebank-len}.

\subsection{English Idioms and Matched Phrase Set}
\label{sec:chip-description}

Previous datasets center around notable bigrams, some of which are compositional and some of which are non-compositional \cite{ramisch-etal-2016-naked, reddy-etal-2011-empirical}. However, there is a positive correlation between bigram frequency and human compositionality scores in these datasets, which means that it is unclear whether models are capturing compositionality or merely frequency effects if they correlate well with the human scores.

Because models are likely more sensitive to surface features of language than humans, we gathered a more controlled set of phrases to compare with human judgments. 

Since non-compositional phrases are somewhat rare, we began with a set of seed idioms and bigrams from previous studies \cite{ jhamtani-etal-2021-investigating,ramisch-etal-2016-naked, reddy-etal-2011-empirical}. We used idioms because they are a common source of non-compositional phrases. Duplicates after lemmatization were removed.

For each idiom, we used Google Syntactic NGrams to find three phrases with an identical part of speech and dependency structure to that idiom, and frequency that was as close as possible relative to others in Syntactic Ngrams \cite{syntactic-ngrams}.%
\footnote{The part of speech/dependency pattern for each idiom was taken to be the most common pattern for that phrase in the dataset}
For example, the idiom "sail under false colors" was matched with "distribute among poor parishioners". More examples can be found in \autoref{tab:idiom-and-match}. An author of this paper inspected the idioms and removed those that were syntactically analyzed incorrectly or offensive. 

\begin{table*}[!htbp]
    \centering
    \small
    \begin{tabular}{cccc}
    \toprule
        Idiom & Matched phrase & Syntactic pattern & Log frequency \\
    \toprule
        Devil's advocate & Baker's town & JJ/dep/2 NN/pobj/0 & 2.398 \\ 
    \midrule
       Act of darkness & Abandonment of institution & NN/dobj/0 IN/prep/1 NN/pobj/2 & 4.304 \\
    \midrule
        School of hard knocks & Field of social studies & NN/pobj/0 IN/prep/1 JJ/amod/4 NNS/pobj/2 & 6.690\\
    \bottomrule
        
    \end{tabular}
    \caption{Examples of idioms with their matched phrases, selected based on having the same syntactic pattern and most similar log frequency in the Syntactic Ngrams dataset. Examples depicted here have the same log frequency. Note that the frequency is based on the most common dependency and constituency pattern found in Syntactic NGrams. Humans were asked to rate each phrase for its compositionality.}
    \label{tab:idiom-and-match}
\end{table*}


\section{Approximating a Composition Function}
\label{sec:comp-fn}

\subsection{Methods}
To approximate the composition functions of models, we extract the \textbf{CLS} and \textbf{AVG} representations from each model on the Treebank dataset. We used 10-fold cross-validation and trained the learned probes on the 90\% training set in each fold. The remaining 10\% were divided into a test set (5\%) and dev set (5\%).\footnote{The learned probes were trained with early stopping on the dev set with a patience of 2 epochs, up to a maximum of 20 epochs. The Adam optimizer was used, with a batch size of 512 and learning rate of 0.512.}

To fairly compare probes, we used minimum description length probing \cite{mdl}.This approximates the length of the online code needed to transmit both the model and data, which is related to the area under the learning curve. Specifically, we recorded average cosine similarity of the predicted vector and actual vector on the test set while varying the size of the training set from 0.005\% to 100\% of the original.\footnote{We look at milestones of 0.005\%, 0.01\%, 0.1\%, 0.5\%, 1\%, 10\% and 100\% specifically. This was because initial experimentation showed that probes tended to converge at or before 10\% of the training data. Models were trained separately (with the same seed and initialization) for each percentage of the training data, and trained until convergence for each data percentage condition. } We compare the AUC of each probe under these conditions to select the most parsimonious approximation for each model.

\subsection{Results}
\begin{figure*}[!ht]
    \centering
    \includegraphics[width=\textwidth]{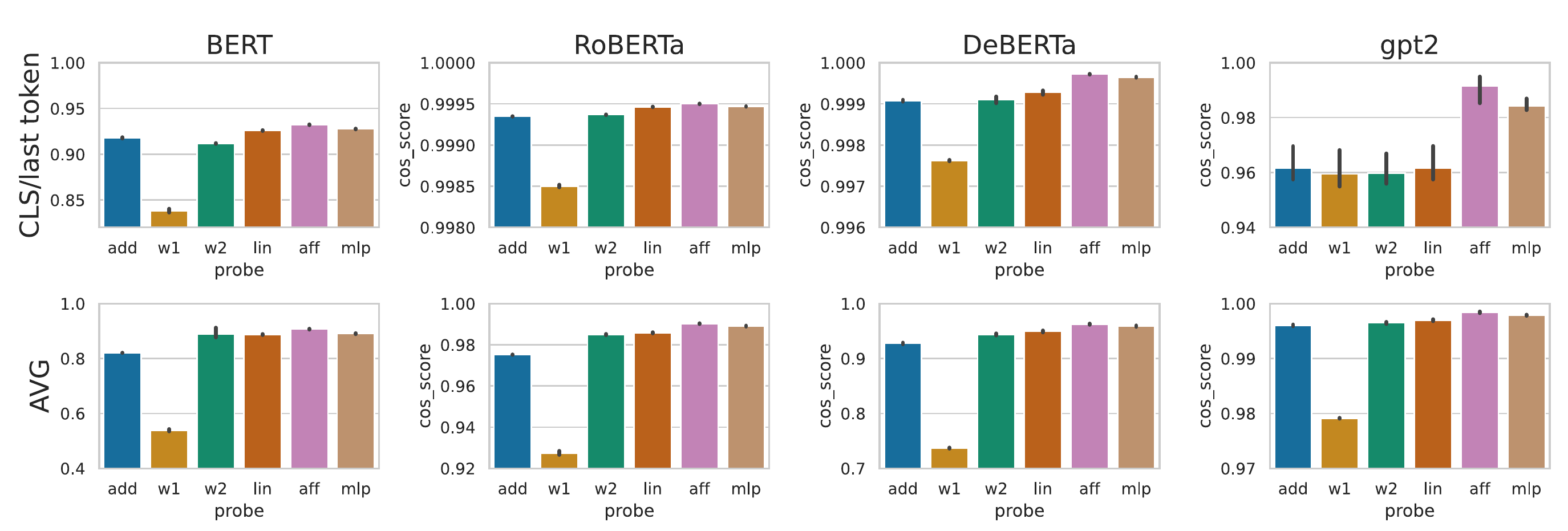}
    \caption{Mean compositionality score (cosine similarity) and standard deviation of each approximative probe across 10 folds. Error bar indicates 95\% CI.}
    \label{fig:probe-perf}
\end{figure*}

We find that \textbf{affine probes} are best able to capture the composition of phrase embeddings from their left and right subphrases. A depiction of probe performance at approximating representations across models and representation types is in \autoref{fig:probe-perf}.  However, we note that scores for most models are very high, due to the anisotropy phenomenon. This describes the tendency for most embeddings from pretrained language models to be clustered in a narrow cone, rather than distributed evenly in all directions \cite{li-etal-2020-sentence, ethayarajh-2019-contextual}. We note that it is true for both word and phrase embeddings. 

Since we are comparing the probes to each other relative to the same anisotropic vectors, this is not necessarily a problem. However, in order to compare each probe's performance compared to chance, we correct for anisotropy using a control task. This task is using the trained probe to predict a random phrase embedding from the set of treebank phrase embeddings for that model, and recording the distance between the compositional probe's prediction and the random embedding. This allows us to calculate an error ratio $\frac{\text{dist}_{\text{probe}}}{\text{dist}_{\text{control}}}$, where $\text{dist}_{probe}$ represents the original average distance from the true representation, and $\text{dist}_{\text{control}}$ is the average distance on the control task. This quantifies how much the probe improves over a random baseline that takes anisotropy into account, where a smaller value is better. These results can be found in \autoref{sec:cos-corrected}. The results without anisotropy correction can be found in \autoref{sec:original-scores}. In most cases, the affine probe still performs the best, so we continue to use it for consistency on all the model and representation types.

We also compare the AUC of training curves for each probe and find that the affine probe remains the best in most cases, except $\text{RoBERTa}_{\text{CLS}}$ and $\text{DeBERTa}_{\text{CLS}}$. Training curves are depicted in \autoref{sec:probe-learning}. AUC values are listed in \autoref{sec:probe-auc}.

Interestingly, there was a trend of the right child being weighted more heavily than the left child, and each model/representation type combination had its own characteristic ratio of the left child to the right child. For instance, in BERT, the weight on the left child was 12, whereas it was 20 for the right child. 

For example, the approximation for the phrase "green eggs and ham" with BERT \texttt{[CLS]} embeddings would be: 
$r_{CLS}(\text{"green eggs and ham"}) = 12 r_{CLS}(\text{"green eggs"}) + 20 r_{CLS}{\text{("and ham")}} + \beta$. 

\section{Examining Compositionality across Phrase Types}
\label{sec:phrase-types}

\subsection{Methods}
Intuitively, we expect the phrases whose representations are close to their predicted representation to be more compositional. We call similarity to the expected representation, $\text{sim}(r(x), \hat{f}(r(a), r(b)))$, the \textit{compositionality score} of a phrase.

We record the mean reconstruction error for each tree type and report the results. In addition to comparing tree types to each other, we also examine the treatment of named entities in \autoref{sec:named-entities}. We examine the relationship between length of a phrase in words and its compositionality score in \autoref{sec:comp-phrase-len}. 

\subsection{Results}

\begin{figure*}[!ht]
    \centering 
    \includegraphics[width=0.75\textwidth]{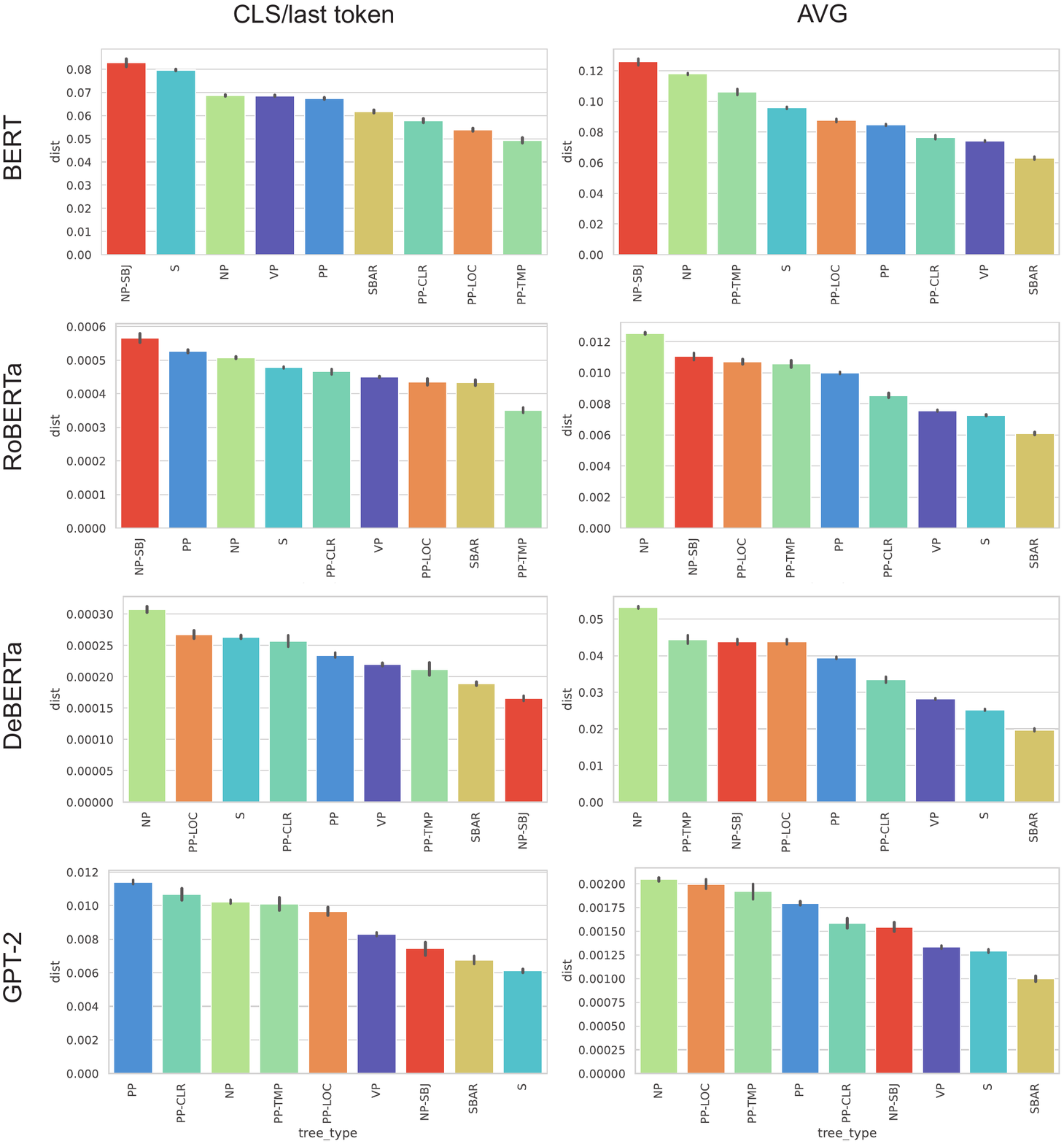}
    \caption{Tree reconstruction error (cosine distance) for each parent phrase type present in the Treebank, ordered from highest mean error to lowest. Based on the affine approximation for each model and representation type. Expanded version with all tree types is presented in \autoref{sec:expanded-phrase-types}.}
    \label{fig:tree-type-dev}
\end{figure*}

There is a significant difference between the mean compositionality score of phrase types. Particularly, the \textbf{AVG} representation assigns a lower compositionality score to NP $\rightarrow$ NNP NNP phrases, which is expected since this phrase type often corresponds to named entities. By contrast, the \textbf{CLS} representation assigns a low compositionality score to NP $\rightarrow$ DT NN, which is unexpected given that such phrases are generally seen as compositional. The reconstruction error for the most common phrase types is shown in \autoref{fig:tree-type-dev}.

Because different phrase types may be treated differently by the model, we examine the relative compositionality of phrases within each phrase type. Examples of the most and least compositional phrases from several phrase types are shown in \autoref{tab:roberta-cls-phrase-types} for $\text{RoBERTa}_{\text{CLS}}$. Patterns vary for model and representation types, but long phrases are generally represented more compositionally.

\begin{table*}[!ht]
    \centering
    \small
    \begin{tabular}{p{0.15\textwidth}p{0.4\textwidth}p{0.4\textwidth}}
        \toprule
        Phrase type & Most compositional &  Least compositional\\
        \toprule
        PP $\rightarrow$ IN NP & ("of", "two perilous day spent among the planters of Attakapas, $\ldots$) & ("of", "September") \\
         & ("of", "the cloth bandoleers that marked the upper part of his body $\ldots$)& ("like", "the Standard \& Poor 's 500") \\
        \midrule
        S $\rightarrow$ NP-SBJ VP & ("him", "to suggest it's the difference between the 'breakup' value $\ldots$) & ("other things", "being more equal")\\
        & ("it", "was doing a brisk business in computer power-surge protectors $\ldots$") & ("less", "is more")\\
        \midrule
        NP $\rightarrow$ NNP NNP & ("M.", "Bluthenzweig") & ("Edward", "Thompson") \\
        & ("Dr.", "Volgelstein") & ("Alexander", "Hamilton") \\
        \bottomrule
    \end{tabular}
    \caption{Phrases rated most and least compositional using $\text{RoBERTa}_{\text{CLS}}$ representations, from several syntactic phrase types. "$\ldots$" indicates that a phrase continues but is too long to display. Long phrases and abbreviated names tend to have a higher compositionality score.}
    \label{tab:roberta-cls-phrase-types}
\end{table*}

\subsubsection{Named Entities}
\label{sec:named-entities}
\begin{figure}[!ht]
    \centering
    \includegraphics[width=\columnwidth]{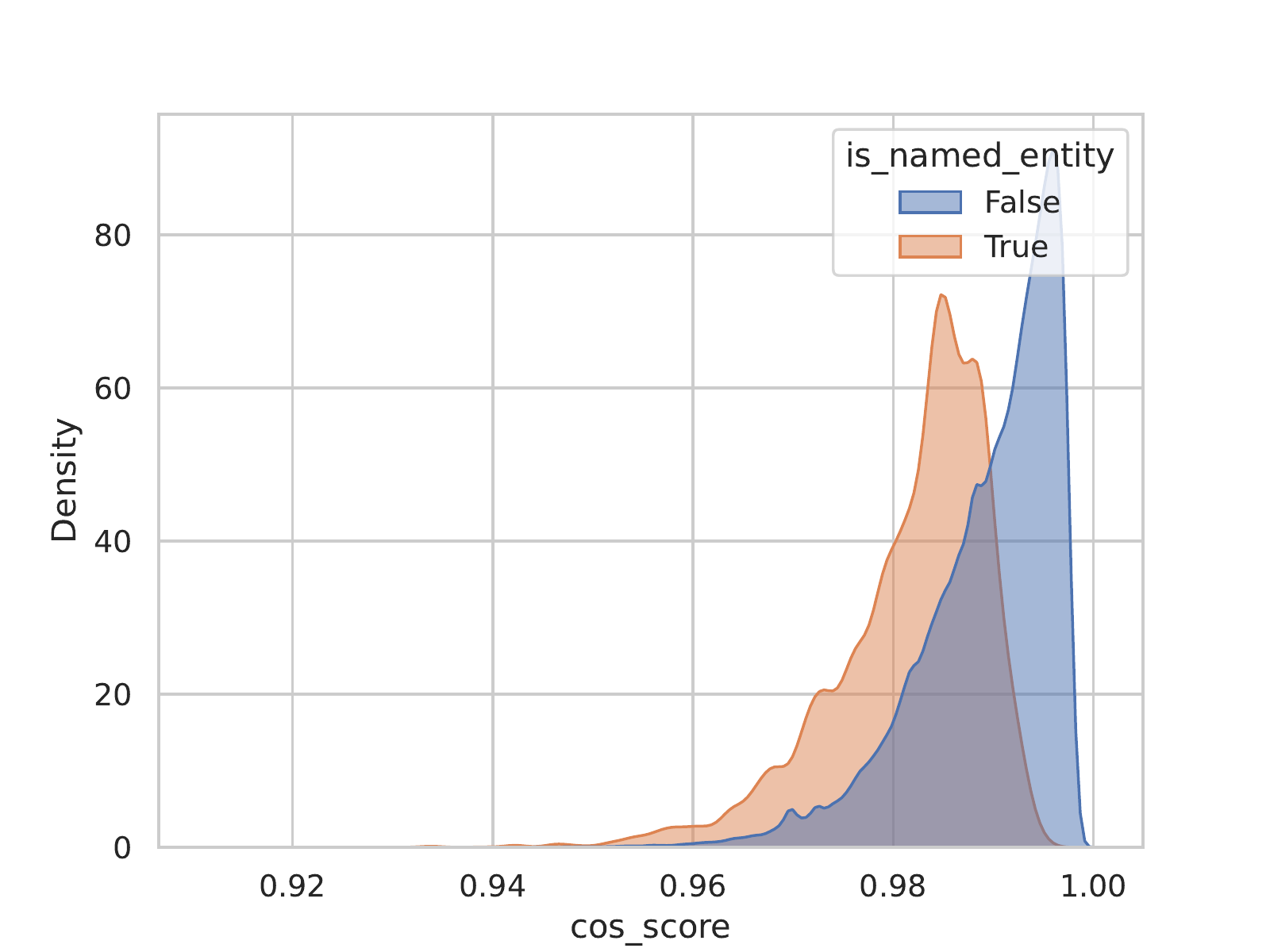}
    \caption{Density plot for compositionality scores of named entities and non-named-entities with $\text{RoBERTa}_{\text{AVG}}$ representations. Higher means more compositional.}
    \label{fig:hist-named-ents}
\end{figure}

\begin{figure}[!ht]
    \centering 
    \includegraphics[width=\columnwidth]{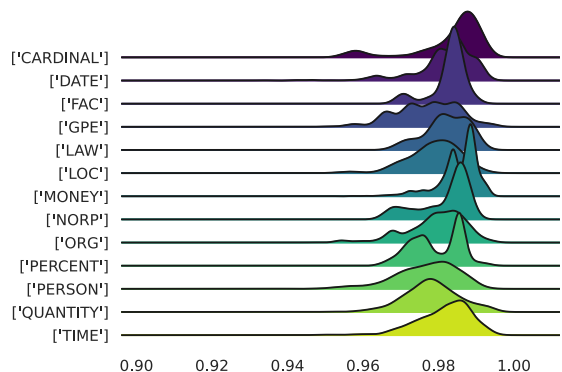}
    \caption{Density plots for compositionality scores of different named entity types with $\text{RoBERTa}_{\text{AVG}}$ representations. Higher means more compositional.}
    \label{fig:named-ent-types}
\end{figure}

We used SpaCy to tag and examine named entities \cite{spacy2}, as they are expected to be less compositional. We find that named entities indeed have a lower compositionality score in all cases except $\text{RoBERTa}_{\text{CLS}}$, indicating that they are correctly represented as less compositional. A representative example is shown in \autoref{fig:hist-named-ents}. Full results can be found in \autoref{sec:named-entities-all}. We break down the compositionality scores of named entities by type and find surprising variation within categories of named entities. For numerical examples, this often depends on the unit used. For example, in $\text{RoBERTa}_{\text{AVG}}$ representations, numbers with "million" and "billion" are grouped together as compositional, whereas numbers with quantifiers ("about", "more than", "some") are grouped together as not compositional. The compositionality score distributions for types of named entities are presented in \autoref{fig:named-ent-types}.

\subsubsection{Examining Compositionality and Phrase Length}
\label{sec:comp-phrase-len}
There is no consistent relationship between phrase length and compositionality score across models and representation types. However, \textbf{CLS} and \textbf{AVG} representations show divergent trends. There is a strong positive correlation between phrase length and compositionality score in the \textbf{AVG} representations, while no consistent trend exists for the \textbf{CLS} representations. This indicates that longer phrases are better approximated as an affine transformation of their subphrase representations. This trend is summarized in \autoref{sec:len-corr}. All correlations are highly significant.

\section{Comparing Compositionality Judgments of Humans and Models}
\label{sec:human-results}

\subsection{Methods}

\subsubsection{Human Annotation}
\label{sec:human-annotation}
Human annotators assigned labels to each phrase in the matched dataset from \autoref{sec:chip-description}: 1 for not compositional, 2 for somewhat compositional, and 3 for fully compositional. They could also decline to answer if they felt that the phrase didn't make sense on its own. Furthermore, they were asked how much each subphrase (left and right) contributed to the final meaning, from 1 for not at all, to 3 for a great deal. The Likert scale of 1-3 was chosen based on analysis of previous compositionality annotation tasks, which found that extreme values of compositionality were the most reliable \cite{comp-judgments}. 

Initially, six English-speaking graduate students were recruited. The six initial annotators all annotated the first 101 examples and the subset of three annotators with the highest agreement who agreed to continue (Krippendorff $\alpha$ = 0.5750) were recruited for the full study, annotating 1001 examples. For the full study, the agreement was higher ($\alpha$ = 0.6633). We took the mean of compositionality judgments to be the final score for phrases. The instructions shown to annotators are in \autoref{sec:annotation}. Examples judgments from an annotator can be found in \autoref{tab:example-annotations}.

\begin{table*}[!ht]
    \centering
    \small
    \begin{tabular}{cccc}
        \toprule
        Phrase &  Idiom & Judgment & Subphrase contribution \\
        \toprule
        Making heavy weather & Yes & 1 - Not compositional & Making: 1 - Not at all \\
        & & & Heavy weather: 2 - Somewhat \\
        \midrule
        Chief part & No & 2 - Somewhat compositional & Chief: 2 - Somewhat \\
        & & & Part: 3 - A great deal \\
        \midrule
        Portrait of Washington & No & 3 - Fully compositional & Portrait: 3 - A great deal \\
        & & & of Washington: 3 - A great deal \\
        \bottomrule
    \end{tabular}
    \caption{Example judgments of one annotator on the pilot set. Annotators were asked to rate each phrase from 1 to 3, where 1 meant not compositional and 3 meant fully compositional. They were also asked how much each subphrase contributed to the meaning.}
    \label{tab:example-annotations}
\end{table*}

\subsubsection{Model Comparison}

To compare human judgments to model compositionality scores, we use the best trained approximative probe for each model and representation type to predict a vector for the full phrase based on its left and right subphrases (taking the probe trained on the first fold). We use cosine similarity to the expected representation as the measure of how compositional a phrase is for a model and representation type. 

We take the Spearman correlation between model compositionality scores and human compositionality judgments and observe differences between human judgments and compositionality scores from model representations.

\subsection{Results}

\subsubsection{Correlation with human judgments}

There is a weak correlation between model and human compositionality scores. The most promising trend is found in RoBERTa, where both \textbf{CLS} and \textbf{AVG} representations have a significant positive correlation with human judgments. Results are in \autoref{tab:human-corr}, with corrected p-values \cite{Holm1979ASS}.

\begin{table}[!ht]
    \centering
    \small
    \begin{tabular}{p{0.15\textwidth}p{0.1\textwidth}p{0.13\textwidth}}
        \toprule
        Model and representation & Spearman $\rho$ & p-val \\
        \toprule
        $\text{BERT}_{\text{CLS}}$ & -0.02308 & 0.9915\\
        $\text{RoBERTa}_{\text{CLS}}$ & 0.1913 & $9.7934 \times 10^{-8}$*\\
        $\text{DeBERTa}_{\text{CLS}}$ & 0.01466 &  0.9915\\
        $\text{GPT-2}_{\text{last}}$ & 0.009428 &  0.02654*\\
        \midrule
        \midrule
        $\text{BERT}_{\text{AVG}}$ & 0.1283 & $8.594 \times 10^{-4}$*\\
        $\text{RoBERTa}_{\text{AVG}}$ & 0.1386 & $2.782 \times 10^{-4}$*\\
        $\text{DeBERTa}_{\text{AVG}}$ & -0.03819 & 0.7792 \\
        $\text{GPT-2}_{\text{AVG}}$ & -0.04598 & 0.6987\\
        \bottomrule
    \end{tabular}
    \caption{Spearman correlation between human judgments of compositionality and compositionality score generated by different model and representation combinations. P-values are corrected for multiple comparisons with the Holm-Bonferroni correction.}
    \label{tab:human-corr}
\end{table}

\subsubsection{Subphrase Contribution Test}

Annotators indicated to what extent they believed each part of the phrase contributed to the final meaning. 
We examined examples in which annotators rated one part of the phrase, for example $a$, as contributing more to the final meaning, and checked how often $d_{cos}(r(x), r(a)) > d_{cos}(r(x), r(b))$. Models do surprisingly poorly at this test, with most performing below chance. Results are presented in \autoref{tab:phrase-test-results}. 
An error analysis on $\text{RoBERTa}_{\text{AVG}}$ indicated that in many cases, errors were due to idiomaticity failures. For example, "noble gas" is a type of gas that was rated as being more similar to "gas" by humans, but "noble" by RoBERTa.\footnote{Similar errors were made for phrases such as "grandfather clock", "as right as rain", "ballpark estimate". A "grandfather clock" is a type of clock, "as right as rain" indicates that something is alright, and a "ballpark estimate" is a rough estimate.}

\begin{table}[!ht]
    \centering
    \small
    \begin{tabular}{cc}
        \toprule
        Model and representation & Subphrase accuracy \\
        \toprule
        $\text{BERT}_{\text{CLS}}$ & 49.71\%\\ 
        $\text{RoBERTa}_{\text{CLS}}$ & 45.91\% \\
        $\text{DeBERTa}_{\text{CLS}}$ & 45.61\% \\
        $\text{GPT-2}_{\text{last}}$ & 43.86\% \\
        \midrule
        \midrule
        $\text{BERT}_{\text{AVG}}$ & 52.92\%\\ 
        $\text{RoBERTa}_{\text{AVG}}$ & 45.03\% \\
        $\text{DeBERTa}_{\text{AVG}}$ & 46.20\% \\
        $\text{GPT-2}_{\text{AVG}}$ & 45.32\% \\
        \midrule
         & Idiomatic accuracy \\
        \toprule
        $\text{BERT}_{\text{CLS}}$ & 45.60\%\\ 
        $\text{RoBERTa}_{\text{CLS}}$ & 60.03\% \\
        $\text{DeBERTa}_{\text{CLS}}$ & 56.67\% \\
        $\text{GPT-2}_{\text{last}}$ & 59.15\% \\
        \midrule
        \midrule
        $\text{BERT}_{\text{AVG}}$ & 57.57\%\\ 
        $\text{RoBERTa}_{\text{AVG}}$ & 58.98\% \\
        $\text{DeBERTa}_{\text{AVG}}$ & 45.77\% \\
        $\text{GPT-2}_{\text{AVG}}$ & 48.42\% \\
        \bottomrule
    \end{tabular}
    \caption{Accuracy of model representations on the subphrase test and idiomaticity test.}
    \label{tab:phrase-test-results}
\end{table}
 
\subsubsection{Idiomaticity Test}
Because idioms were matched with non-idiomatic expressions, we tested for correctly identifying the idioms. We limited the analysis to pairs where the idiomatic expression was rated as less compositional than the matched expression. Results are shown in \autoref{tab:phrase-test-results}. Results are better than the subphrase contribution test, but models do not achieve good results, the best performing representation being $\text{RoBERTa}_{\text{CLS}}$.

\subsubsection{Correlations with Other Factors}

We examine correlations of model and human compositionality scores with the frequency and length of the phrase in words. As noted before, there is a strong correlation between length and compositionality score in models but not in human results. Results are in \autoref{sec:corr-freq-len-human}. A comparison of phrases rated as most and least compositional by humans, as well as RoBERTa, is presented in \autoref{tab:comp-judgments-examples}.

\begin{table}[!ht]
    \centering
    \small
    \begin{tabular}{p{0.09\textwidth}p{0.15\textwidth}p{0.15\textwidth}}
        \toprule
        Model \& representation & Most compositional & Least compositional \\
        \toprule
        Human & "population growth" & "gravy train"\\
        & "few weeks away" & "shrinking violet" \\
        & "railroad monopoly" & "revolving door syndrome" \\
        \midrule
        $\text{RoBERTa}_{\text{CLS}}$ & "two small sticks" & "worse than none" \\
        & "dark glass bottle" & "cases apart" \\
        & "annual music festival" & "arch'd eyebrow" \\
        \midrule
        $\text{RoBERTa}_{\text{AVG}}$ & "look with open eyes" & "advertisement revenue" \\
        & "be of equal importance" & "taking it upon oneself" \\
        & "come after breakfast" & "all paces" \\
        \bottomrule
    \end{tabular}
    \caption{Most and least compositional phrases in CHIP by human judgments and RoBERTa compositionality scores. Human scores are the average of 3 annotators.}
    \label{tab:comp-judgments-examples}
\end{table}

\section{Related work}

\subsection{Background on Compositionality}

Compositionality has been debated in the philosophy of language, with opposing views \cite{herbelot2020how}: the \textit{bottom-up} view that the meaning of a larger phrase is a function of the meaning of its parts \cite{cresswell}, and the \textit{top-down} view that smaller parts only have meaning as a function of the larger phrase \cite{fodorlepore}. It is likely that there is a blend of bottom-up and top-down processing corresponding to compositional and non-compositional phrases respectively \cite{dankers2021-paradox}. 

Hupkes et al. have proposed several compositionality tests based on previous interpretations: \cite{hupkes-review}. We focus on localism, corresponding to the bottom-up view. 

\subsection{Other Definitions of Compositionality}

Other works do other tests for compositionality, notably substitutivity \cite{hupkes-review}. Evidence suggests that models may be unable to modulate the bottom-up and top-down processing of phrases \cite{dankers-etal-2022-transformer, dankers2021-paradox}. Substitutivity effects appear to not be represented well \cite{garcia-etal-2021-probing, Yu2020AssessingPR}. This indicates that phrases are not being composed as expected and motivates our study of how local composition is carried out in these models, and which types of phrase are processed top-down and bottom-up.




\subsection{Studies of Localism}

Previous studies of local composition focus on bigrams, particularly adjective-noun and noun-noun bigrams \cite{nandakumar-etal-2019-well, cordeiro-etal-2019-unsupervised, salehi-etal-2015-word, reddy-etal-2011-empirical, mitchell-lapata}. However, many of these studies assume an additive composition function or only fit a composition function on the bigrams in their datasets. 

A study finds some evidence for successful local composition in the case of mathematical expressions, but used a constrained test set on a domain that is expected to be perfectly locally compositional \cite{Russin2021CompositionalPE}.

\subsection{Approximating LM Representations}

There has been recent interest in understanding the compositionality of continuous representations generated by neural models \cite{https://doi.org/10.48550/arxiv.2205.01128}. LM representations have been approximated as the output of explicitly compositional networks based on tensor products \cite{tpdn2, rnn-tpdn, role-network}. These are typically evaluated based on compositional domains, such as the SCAN dataset \cite{Lake2017StillNS}. 

Previous work on the geometry of word embeddings within a sentence shows that language models can encode hierarchical structure \cite{bert-geometry, manning2020, jawahar-etal-2019-bert}. However, it is an open question as to why LMs do not tend to generalize well compositionally \cite{Lake2017StillNS, Keysers2020MeasuringCG}. 

\section{Conclusion}

We analyze the compositionality of representations from several language models and find that there is an effective affine approximation in terms of a phrase's syntactic children for many phrases. Although LM representations may be surprisingly predictable, we find that human compositionality judgments do not align well with how LM representations are structured.

In this work, we study the representations produced after extensive training. However, the consistency of several trends we observed suggests that there may be theoretical reasons why LM representations are structured in certain ways. Future work could investigate the evolution of compositionality through training, or motivate methods that would allow LMs to achieve improved compositional generalization while representing non-compositionality. 

\section*{Acknowledgments}

Thank you to Amanda Bertsch, Ting-Rui Chiang, Varun Gangal, Perez Ogayo, and Zora Wang for participating in compositionality annotations. This work was supported in part by a CMU Presidential Fellowship to the first author, and the Tang Family AI Innovation Fund.

\section*{Limitations}

One limitation of this work is that it was conducted on a relatively small set of language models trained on English, and the diversity of patterns within even this set of language models and representation types is great. However, we note that the experiments can be easily repeated for any language that has a treebank or good-quality syntactic parsers. A related limitation is that these analyses are dependent on what we take to be the "child" constituents of a parent phrase. It may be harder to examine compositionality for languages that differ substantially from English, or that cannot be easily parsed using existing tools. 

Although we try to carefully catalog behaviour observed on natural language phrases, it is likely that smaller-scale experiments providing a more mechanistic understanding of model behaviour would be easier to parse for readers. Although this would be ideal, we leave this for future work, as our main goal was to examine how language models represent phrases considered to be compositional and non-compositional in natural language.

Another limitation is that although we diagnose a problem in language models, we do not provide a clear avenue to fix it. Further work could be done to understand what data distributions or training methods encourage model representations to be more aligned with human judgments. Additionally, although compositionality is linguistically important, more effort could be put towards understanding the downstream tasks for which it is more important. For instance, there could be clear issues in machine translation if non-compositional phrases are not represented properly, but these phrases may not be important in other areas such as instruction following or code generation.

\section*{Ethics Statement}

\subsection*{Potential Risks and Impacts}
Although we aim to document compositionality effects in English, we acknowledge that this perpetuates the problem of English being the dominant language in NLP research. It is possible that conclusions here do not hold for other languages, and further work is needed to understand whether these conclusions transfer.

Additionally, although we tried to filter out offensive idioms from \textbf{CHIP}, this was based on one person's best judgment, and it is possible that some of the terms in the dataset may be offensive to some people. Overall, phrases in the dataset tend to be benign, but some idioms are meant to have a perjorative meaning. 

\subsection*{Computational Infrastructure and Computing Budget}
To run our computational experiments, we made use of a shared compute cluster. We used approximately 100 GPU hours to run experiments, mainly due to running results for different language models and representation types. We did not have any computational budget besides that already used to maintain the cluster.


\bibliography{anthology,custom}
\bibliographystyle{acl_natbib}

\appendix
\clearpage

\section{Treebank dataset tree types}
\label{sec:treebank-tree-types}

Due to space constraints, we only show the top 20 tree types. This can be found in \autoref{tab:tree-types}.

\begin{table*}[!ht]
    \centering
    \small
    \begin{tabular}{ccc}
        \toprule
        Tree type & Count & Example\\
        \toprule
        PP $\rightarrow$ IN NP & 77716 & ((in) (american romance))\\
        S $\rightarrow$ NP-SBJ VP & 62948 & ((he) (said simultaneously, "i wish they were emeralds")) \\
        NP $\rightarrow$ DT NN & 40876 & ((the) (way)) \\
        NP $\rightarrow$ NP PP & 35743 & ((the temporal organization) (of the dance)) \\
        S $\rightarrow$ NP-SBJ S|<VP-.> & 24467 & ((the partners) (said they already hold 15 \% of all shares outstanding.))\\
        VP $\rightarrow$ TO VP & 21833 & ((to) (be the enemy))\\
        PP-LOC $\rightarrow$ IN NP & 18005 & ((in) (the marketplace)) \\
        NP $\rightarrow$ DT NP|<JJ-NN> & 14898 & ((a) (professional linguist))\\
        VP $\rightarrow$ MD VP & 13575 & ((could) (make up his mind)) \\
        VP $\rightarrow$ VB NP & 11838 & ((evaluate) (the progress of therapy)) \\
        PP-TMP $\rightarrow$ IN NP & 11032 & ((for) (almost a year))\\ 
        PP-CLR $\rightarrow$ IN NP & 10054 & ((from) (the most sympathetic angle)) \\
        NP $\rightarrow$ NNP NNP & 9863 & ((honolulu) (harbor)) \\
        NP $\rightarrow$ JJ NNS & 9477 & ((recent) (years)) \\
        VP $\rightarrow$ VBD VP & 8356 & ((was) (salted)) \\
        SBAR $\rightarrow$ WHNP-1 S & 8332 & ((what) (to look for)) \\
        SBAR $\rightarrow$ IN S & 7848 & ((that) (it exceeds the company 's annual sales and its market capitalization)) \\
        NP-SBJ $\rightarrow$ DT NN & 7600 & ((the) (rebound)) \\
        S $\rightarrow$ NP-SBJ-1 VP & 7486 & (draperies) (could be designed to serve structural purposes) \\
        NP $\rightarrow$ NP SBAR & 7317 & ((the " culture shock ") (they might encounter in remote overseas posts)) \\
        \bottomrule
    \end{tabular}
    \caption{Counts of the top 20 grammatical tree types found in the WSJ and Brown sections of the Penn Treebank, with some examples given.}
    \label{tab:tree-types}
\end{table*}

\section{Treebank dataset phrase lengths}
\label{sec:treebank-len}

\begin{figure}[!ht]
    \centering
    \includegraphics[width=\columnwidth]{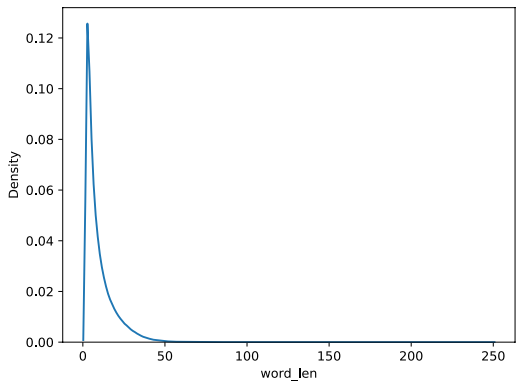}
    \caption{Length distribution of phrases mined from the treebank, in number of words. The modal length was 3 words, followed closely by 2 words. Few phrases contained more than 50 words.}
    \label{fig:treebank-word-len}
\end{figure}

\section{Probe learning curves}

\begin{figure*}[!ht]
    \centering
    \includegraphics[width=\textwidth]{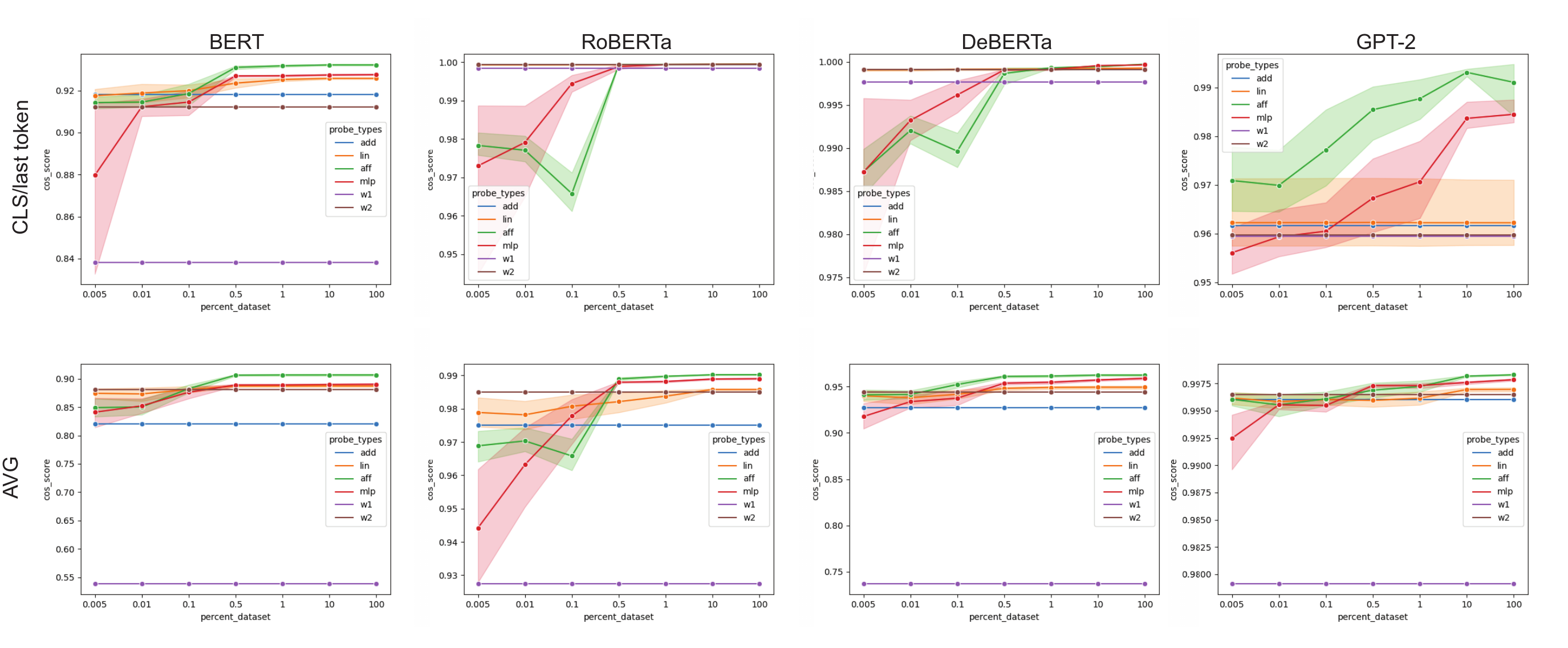}
    \caption{Learning curves of approximative probes trained on differing percentages of train data.}
    \label{fig:probe-learning}
\end{figure*}

Learning curves of the approximative probes (across 10 folds) are shown in \autoref{fig:probe-learning}.

\label{sec:probe-learning}

\section{Length Correlation}
\label{sec:len-corr}

The correlations of the phrase length (in words) and compositionality scores in Treebank are shown in \autoref{tab:len-corr}.

\begin{table}[!ht]
    \centering
    \small
    \begin{tabular}{ccc}
    \toprule
        Model and representation & Spearman $\rho$ & p-val\\
    \toprule
        $\text{BERT}_{\text{CLS}}$ & -0.0700 & 0.0 \\
        $\text{RoBERTa}_{\text{CLS}}$ & 0.1659 & 0.0\\
        $\text{DeBERTa}_{\text{CLS}}$ & 0.1166 & 0.0 \\
        \midrule
        \midrule
        $\text{BERT}_{\text{AVG}}$ & 0.7143 & 0.0 \\
        $\text{RoBERTa}_{\text{AVG}}$ & 0.7086 & 0.0\\
        $\text{DeBERTa}_{\text{AVG}}$ & 0.7866 & 0.0 \\
        \bottomrule
    \end{tabular}
    \caption{Spearman $\rho$ correlation between phrase length (in words) and compositionality score in the treebank.}
    \label{tab:len-corr}
\end{table}

\section{Error ratio of probes}
\label{sec:cos-corrected}
\begin{table}[H]
    \centering
    \small
    \begin{tabular}{ccc}
        \toprule
        Model/representation & Probe & Mean err. ratio ($\downarrow$) \\
        \toprule
        $\text{BERT}_{\text{CLS}}$ & ADD & 0.4668 \\
        & W1 & 0.7806 \\
        & W2 & 0.3903 \\
        & LIN & 0.3940  \\
        & AFF & 0.3908  \\
        & \textbf{MLP} & \textbf{0.3830} \\ \midrule
        $\text{RoBERTa}_{\text{CLS}}$ & ADD & 0.4152 \\
        & W1 & 0.7946 \\
        & \textbf{W2} & \textbf{0.2980} \\
        & LIN & 0.3063  \\
        & AFF & 0.3013  \\
        & MLP & 0.3065 \\ \midrule
        $\text{DeBERTa}_{\text{CLS}}$ & ADD & 0.7577 \\
        & \textbf{W1} & \textbf{0.4661} \\
        & W2 & 0.7090 \\
        & LIN & 0.6777  \\
        & AFF & 0.9373  \\
        & MLP & 0.5856 \\ \midrule
        $\text{GPT-2}_{\text{last}}$ & ADD & 0.4668 \\
        & W1 & 0.7806 \\
        & \textbf{W2} & \textbf{0.3903} \\
        & LIN & 0.3940  \\
        & AFF & 0.3908  \\
        & MLP & 0.3830 \\ \midrule \midrule
        $\text{BERT}_{\text{AVG}}$ & ADD & 0.3873 \\
        & W1 & 0.8060 \\
        & W2 & 0.2167 \\
        & LIN & 0.2327  \\
        & \textbf{AFF} & \textbf{0.2098}  \\
        & MLP & 0.2283 \\ \midrule
        $\text{RoBERTa}_{\text{AVG}}$ & ADD & 0.4504 \\
        & W1 & 0.8422 \\
        & W2 & 0.2431 \\
        & LIN & 0.2471  \\
        & \textbf{AFF} & \textbf{0.2095}  \\
        & MLP & 0.2181 \\ \midrule
        $\text{DeBERTa}_{\text{AVG}}$ & ADD & 0.4472 \\
        & W1 & 0.8886 \\
        & W2 & 0.3202 \\
        & LIN & 0.3143  \\
        & \textbf{AFF} & \textbf{0.3044}  \\
        & MLP & 0.2952 \\ \midrule
        $\text{GPT-2}_{\text{AVG}}$ & ADD & 0.5013 \\
        & W1 & 0.9074 \\
        & W2 & 0.4226 \\
        & LIN & 0.4041  \\
        & \textbf{AFF} & \textbf{0.3475}  \\
        & MLP & 0.3554 \\ \bottomrule
    \end{tabular}
    \caption{Error ratio ($\frac{\text{dist}_{\text{probe}}}{\text{dist}_{\text{control}}}$) for probes trained to predict representations from different model types. Mean across 10 folds.}
    \label{tab:cos-corrected}
\end{table}

\clearpage

\section{Annotation setup and instructions}
\label{sec:annotation}

Annotators were recruited from a population of graduate students. Initially, 6 annotators completed the pilot experiment, which consisted of 101 examples. The subset of three annotators with highest agreement was asked if they would like to complete the full study. One annotator in the highest-agreement group could not continue to the full study, so this annotator was excluded, and the next group with highest agreement was chosen. The agreement values in \autoref{sec:human-annotation} are for the final group of annotators chosen.

\begin{figure}[H]
    \centering
    \includegraphics[width=\columnwidth]{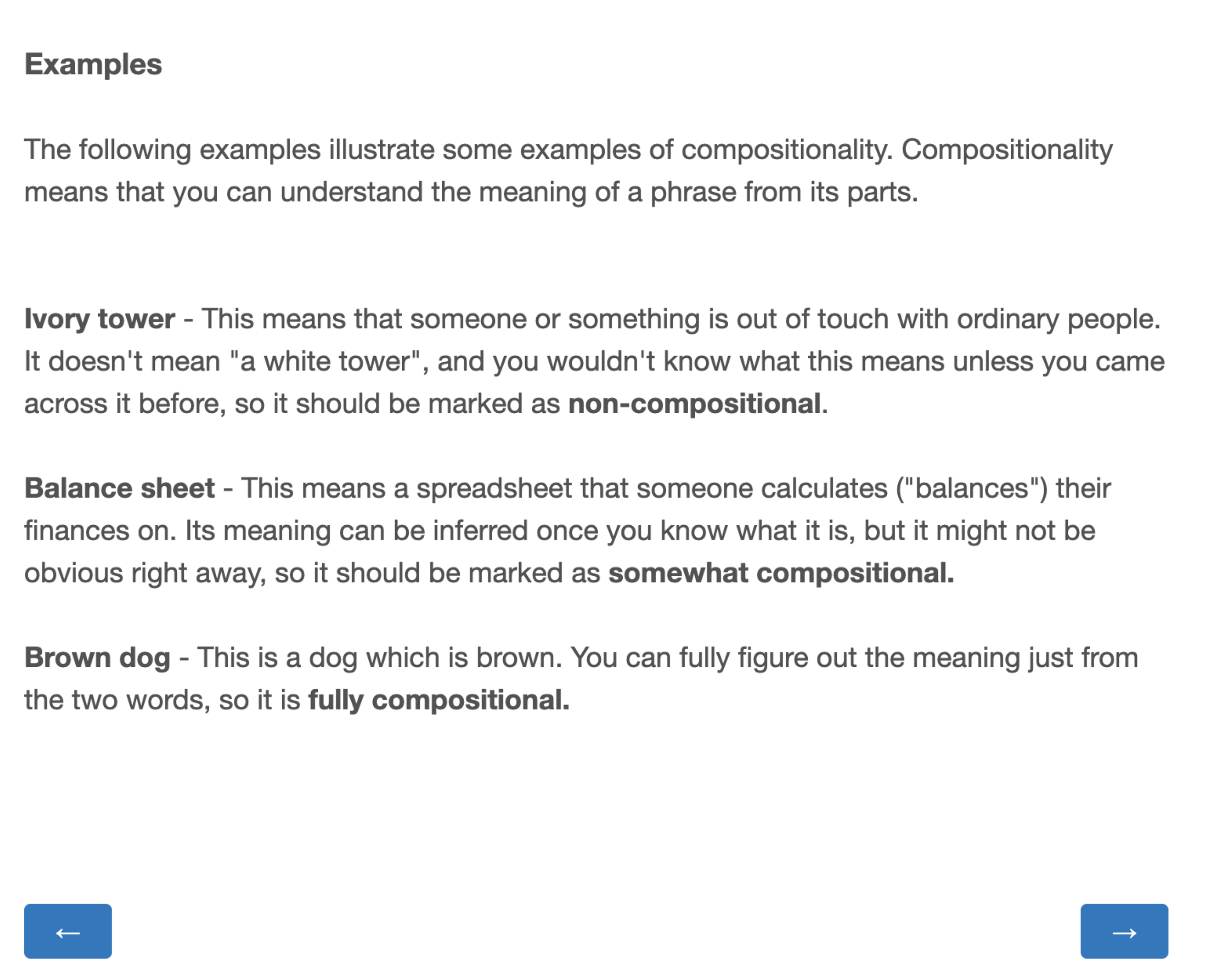}
    \caption{Examples of compositionality judgments shown to annotators}
    \label{fig:qualtrics-examples}
\end{figure}

\begin{figure}[H]
    \centering
    \includegraphics[width=\columnwidth]{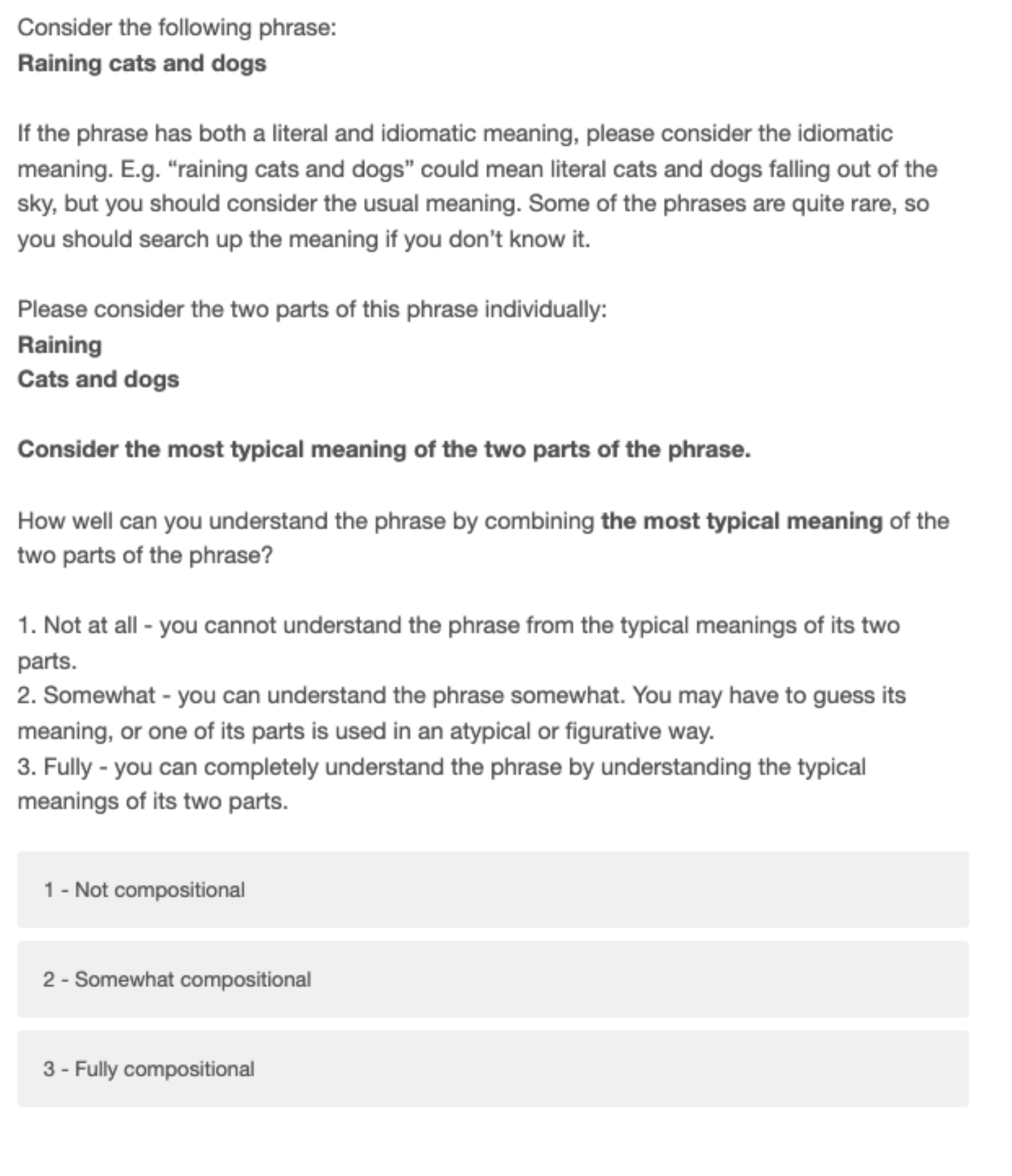}
    \caption{First page of annotation interface for a practice phrase}
    \label{fig:annotation-interface-1}
\end{figure}

\begin{figure}[H]
    \centering
    \includegraphics[width=\columnwidth]{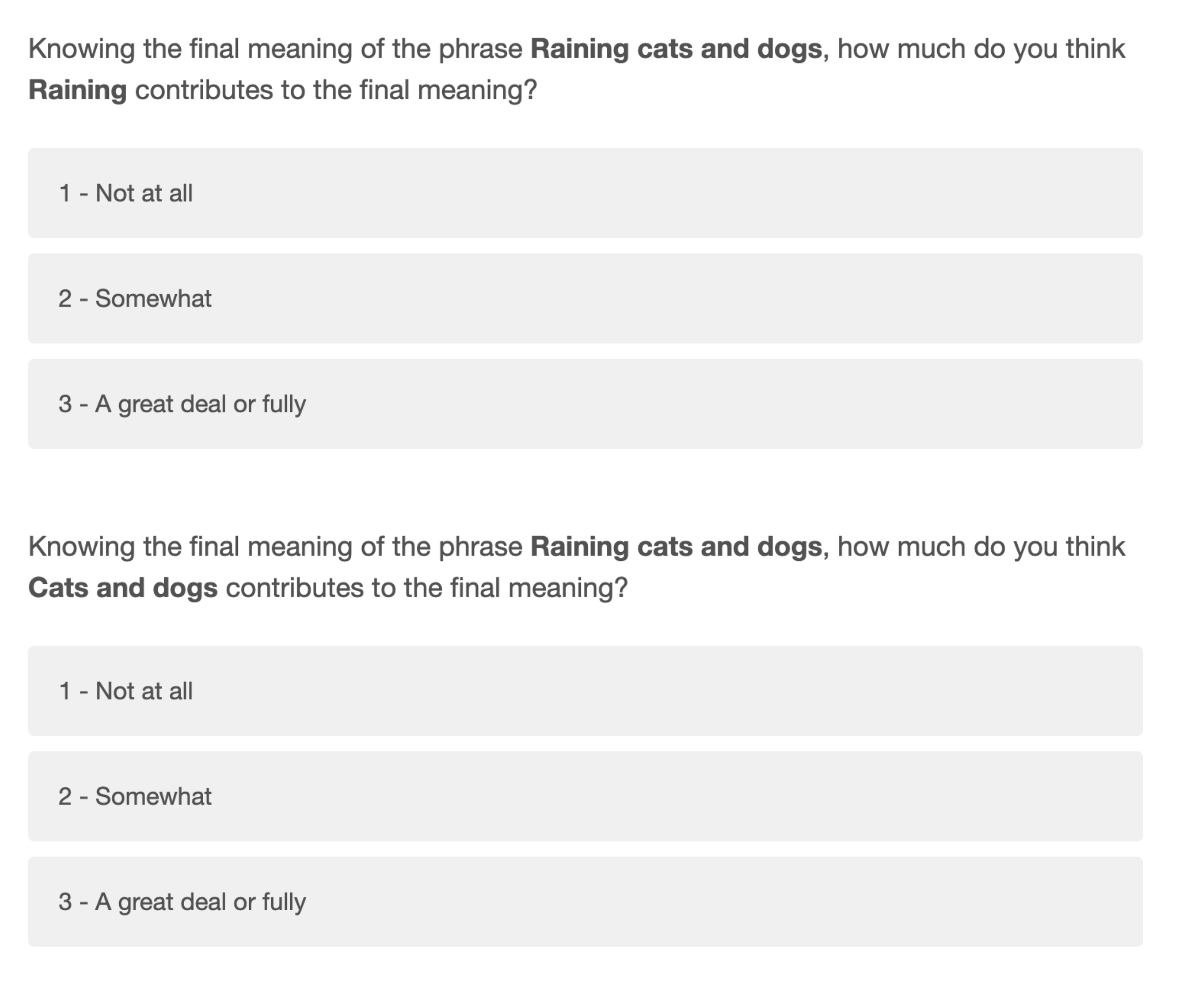}
    \caption{Second page of annotation interface for a practice phrase}
    \label{fig:annotation-interface-2}
\end{figure}

The experiment was implemented on the Qualtrics platform, and participants were first presented with a consent form, linking to more background information on the study, and informing them that their participation was entirely voluntary. After agreeing to the terms, participants were shown some examples and went through 3 practice questions. The example given are shown in \autoref{fig:qualtrics-examples}, and the annotation interface is shown in \autoref{fig:annotation-interface-1} and \autoref{fig:annotation-interface-2}. After completing the practice section, annotators began annotating the real examples, which followed the same interface as the practice examples.

Annotators were all located in the United States, paid approximately \$15 per hour for their work.

\clearpage
\section{Compositionality scores without anisotropy correction}
\label{sec:original-scores}

The raw compositionality scores can be found in \autoref{tab:cos-uncorrected}.

\begin{table*}[!htbp]
    \centering
    \small
    \begin{tabular}{cccc}
    \toprule
        Model and representation & Probe & Mean reconstruction score & Standard dev. \\
    \toprule
        $\text{BERT}_{\text{CLS}}$ &  ADD & 0.9178 & 0.001159\\
        & W1 & 0.8382 & 0.003599 \\
        & W2 & 0.9117 & 0.0007133 \\
        & LIN & 0.9258 & 0.0002285 \\
        & \textbf{AFF} & \textbf{0.9322} & 0.0002033 \\
        & MLP & 0.9276 & 0.0002108 \\
        \midrule
        $\text{RoBERTa}_{\text{CLS}}$ & ADD & 0.99935 & 3.895 $\times 10^{-6}$\\
        & W1 & 0.99850 & 2.612 $\times 10^{-5}$ \\
        & W2 & 0.99937 & 6.866 $\times 10^{-6}$ \\
        & LIN & 0.99946 & 4.735 $\times 10^{-6}$ \\
        & \textbf{AFF} & \textbf{0.99950} & 6.093 $\times 10^{-6}$ \\
        & MLP & 0.99947 & 4.719 $\times 10^{-6}$ \\
        \midrule
        $\text{DeBERTa}_{\text{CLS}}$ & ADD & 0.99908 & 4.070 $\times 10^{-5}$ \\
        & W1 & 0.99762 & 2.900 $\times 10^{-5}$ \\
        & W2 & 0.99911 & 1.399 $\times 10^{-4}$ \\
        & LIN & 0.99928 & 8.963 $\times 10^{-5}$ \\
        & \textbf{AFF} & \textbf{0.99972} & 1.542 $\times 10^{-5}$ \\
        & MLP & 0.99965 & 2.323 $\times 10^{-5}$ \\
        \midrule
        \midrule
        $\text{BERT}_{\text{AVG}}$ & ADD & 0.8205 & 0.0003836 \\
        & W1 & 0.5383 & 0.007471 \\
        & W2 & 0.8893 & 0.03071 \\
        & LIN & 0.8873 & 0.003071 \\
        & \textbf{AFF} & \textbf{0.9069} & 0.002566 \\
        & MLP & 0.8904 & 0.002988 \\
        \midrule
        $\text{RoBERTa}_{\text{AVG}}$ & ADD & 0.9752 & 0.0001306 \\
        & W1 & 0.9274 & 0.001695 \\
        & W2 & 0.9850 & 0.0005092 \\
        & LIN & 0.9858 & 0.0004573 \\
        & \textbf{AFF} & \textbf{0.9902} & 0.0003076 \\
        & MLP & 0.9890 & 0.0003981 \\
        \midrule
        $\text{DeBERTa}_{\text{AVG}}$ & ADD & 0.9275 & 0.002634 \\
        & W1 & 0.7368 & 0.001575 \\
        & W2 & 0.9438 & 0.003321 \\
        & LIN & 0.9493 & 0.003036 \\
        & \textbf{AFF} & \textbf{0.9625} & 0.001814 \\
        & MLP & 0.9590 & 0.002145 \\
        \midrule
        $\text{GPT-2}_{\text{AVG}}$ & ADD & 0.9960 & 0.0002833 \\
        & W1 & 0.9791 & 0.0001214 \\
        & W2 & 0.9965 & 0.0003359 \\
        & LIN & 0.9970 & 0.0003036 \\
        & \textbf{AFF} & \textbf{0.9984} & 0.0002617 \\
        & MLP & 0.9979 & 0.0001634 \\
        \bottomrule
    \end{tabular}
    \caption{Mean reconstruction score (cosine similarity) and standard deviation of each approximative probe across 10 folds. Not corrected for anisotropy in each representation/model type.}
    \label{tab:cos-uncorrected}
\end{table*}


\section{AUC of approximative probes}
\label{sec:probe-auc}

\begin{table}[H]
    \centering
    \small
    \begin{tabular}{ccc}
    \toprule
    Model and representation & probe & AUC \\
    \toprule
    $\text{BERT}_{\text{CLS}}$ &  ADD & 91.80 \\
     & W1 & 83.82 \\
     & W2 & 91.20 \\
     & LIN & 92.57 \\
     & \textbf{AFF} & \textbf{93.20} \\
     & MLP & 92.74 \\
    \midrule
    $\text{RoBERTa}_{\text{CLS}}$ &  ADD & 99.93 \\
     & W1 & 99.84 \\
     & W2 & 99.93 \\
     & LIN & 99.94 \\
     & AFF & 99.93 \\
     & \textbf{MLP} & \textbf{99.94} \\
    \midrule
    $\text{DeBERTa}_{\text{CLS}}$ &  ADD & 99.90 \\
     & W1 & 99.75 \\
     & W2 & 99.90 \\
     & LIN & 99.92 \\
     & AFF & 99.94 \\
     & \textbf{MLP} & \textbf{99.95}\\
     \midrule
     & \textbf{MLP} & \textbf{99.95} \\
    $\text{GPT-2}_{\text{last}}$ & ADD & 96.16 \\
    & W1 & 95.94 \\
    & W2 & 95.97 \\
    & LIN & 96.21 \\
    & \textbf{AFF} & \textbf{99.18} \\
    & MLP & 98.32 \\
    \midrule
    \midrule
    $\text{BERT}_{\text{AVG}}$ &  ADD & 82.04 \\
     & W1 & 53.83 \\
     & W2 & 88.10 \\
     & LIN & 88.68 \\
     & \textbf{AFF} & \textbf{90.63} \\
     & MLP & 88.96 \\
    \midrule
    $\text{RoBERTa}_{\text{AVG}}$ &  ADD & 97.51 \\
     & W1 & 92.73 \\
     & W2 & 98.49 \\
     & LIN & 98.56 \\
     & \textbf{AFF} & \textbf{99.00} \\
     & MLP & 98.88 \\
    \midrule
    $\text{DeBERTa}_{\text{AVG}}$ &  ADD & 92.74 \\
     & W1 & 73.67 \\
     & W2 & 94.38 \\
     & LIN & 94.89 \\
     & \textbf{AFF} & \textbf{96.21} \\
     & MLP & 95.75 \\
    \midrule
    $\text{GPT-2}_{\text{AVG}}$ & ADD & 99.60 \\
    & W1 & 97.90 \\
    & W2 & 99.64 \\
    & LIN & 99.69 \\
    & \textbf{AFF} & \textbf{99.81} \\
    & MLP & 99.76 \\
    \bottomrule
    \end{tabular}
    \caption{AUC scores for probes trained on various percentages of the training set.}
    \label{tab:probe-auc}
\end{table}

\section{Mean deviation of phrase types by tree type}
\label{sec:expanded-phrase-types}

\begin{figure*}[!htbp]
    \centering 
    \includegraphics[width=\textwidth]{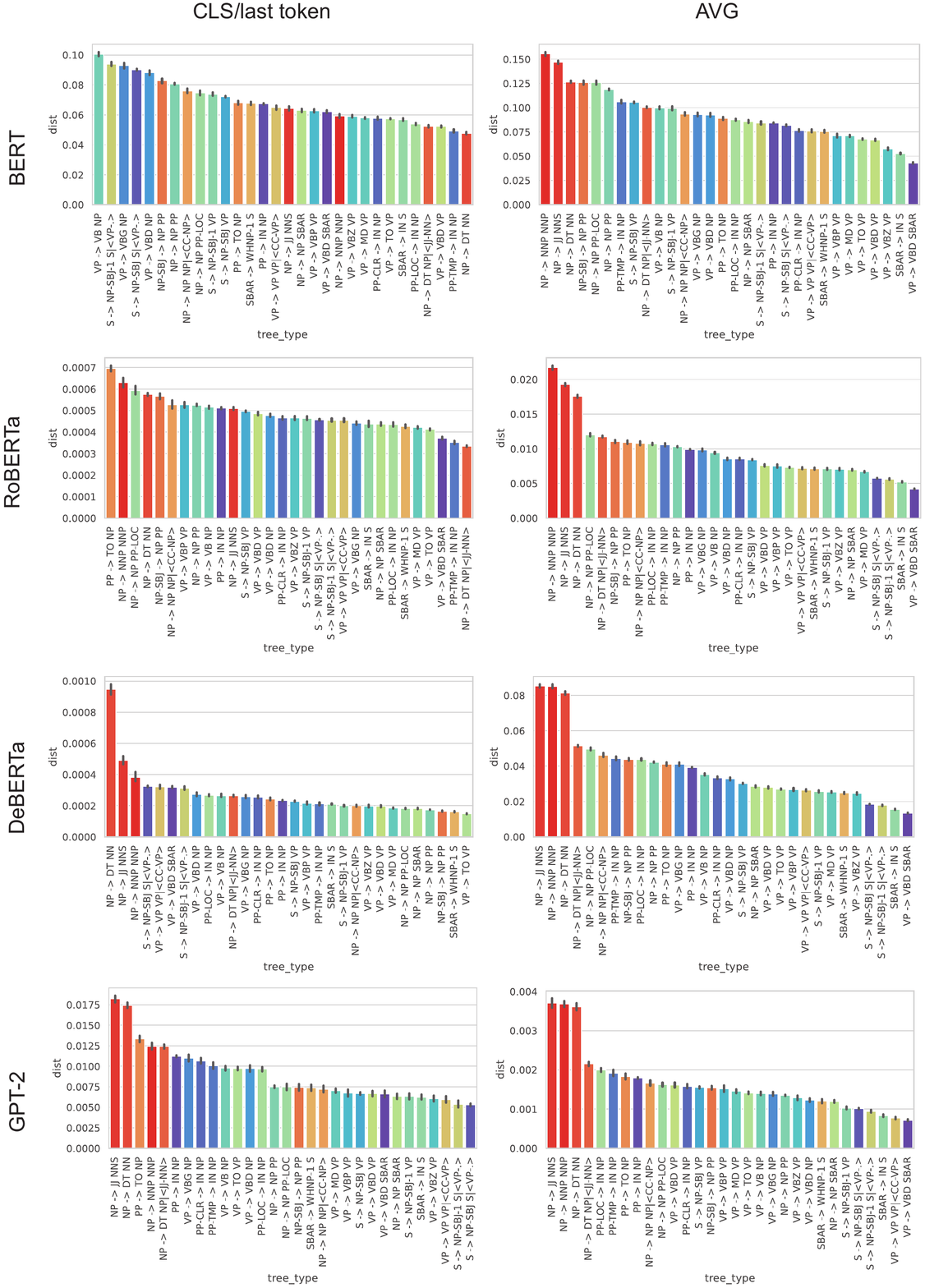}
    \caption{Mean deviation from predicted representation across full tree types.}
    \label{fig:deviation-all-tree-types}
\end{figure*}

The mean deviation of the most common tree types can be found in \autoref{fig:deviation-all-tree-types}.

\section{Further named entity results}
\label{sec:named-entities-all}

\begin{figure*}[!ht]
    \centering
    \includegraphics[width=0.8\textwidth]{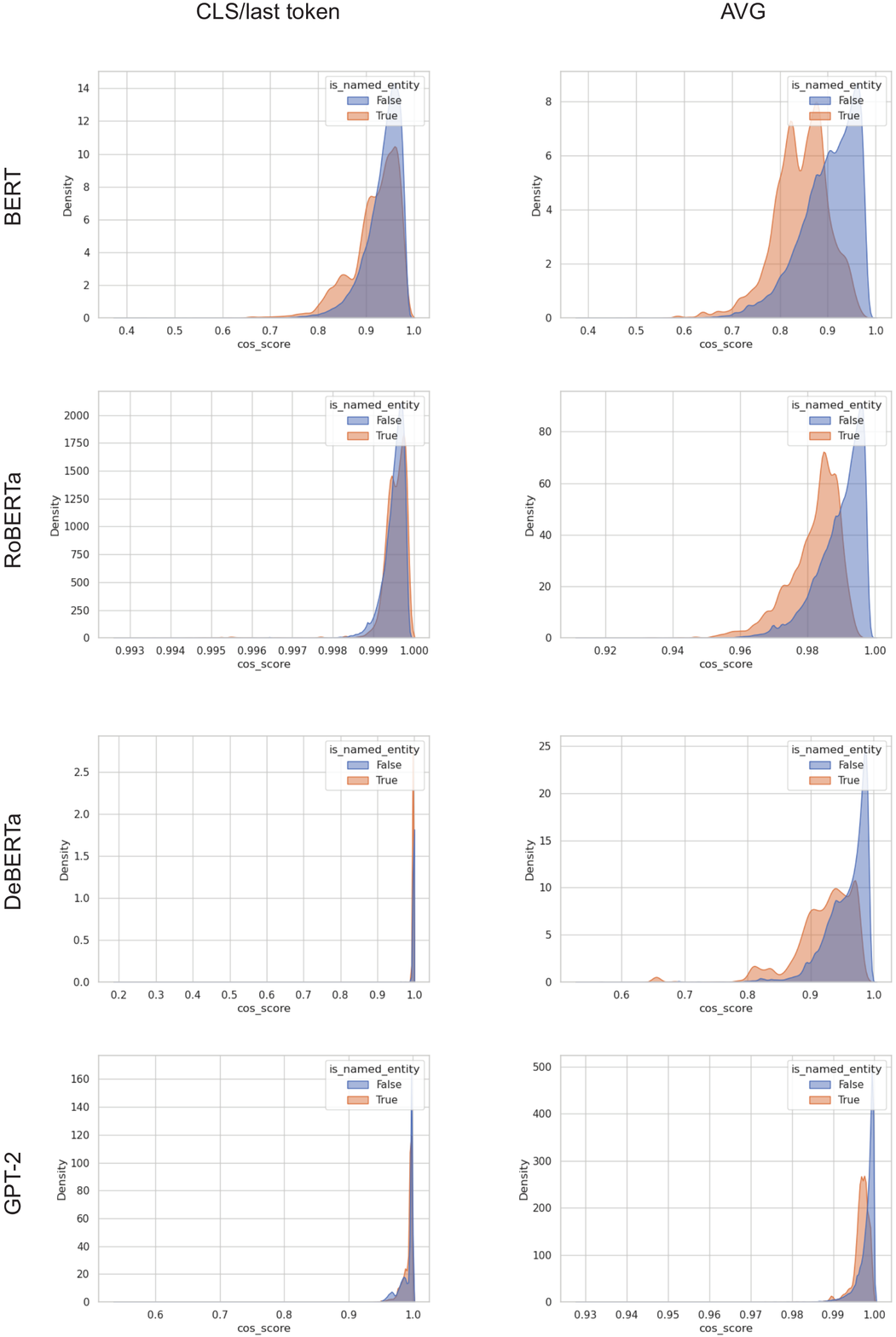}
    \caption{Distributions of compositionality score for named entities and non-named entities across model types and representation types. The AVG representation matches the intuition that named entities are usually less semantically compositional, as they point to an entity in the real world that may not relate to their name.}
    \label{fig:ner-all}
\end{figure*}

\begin{figure*}[!ht]
    \centering
    \includegraphics[width=0.8\textwidth]{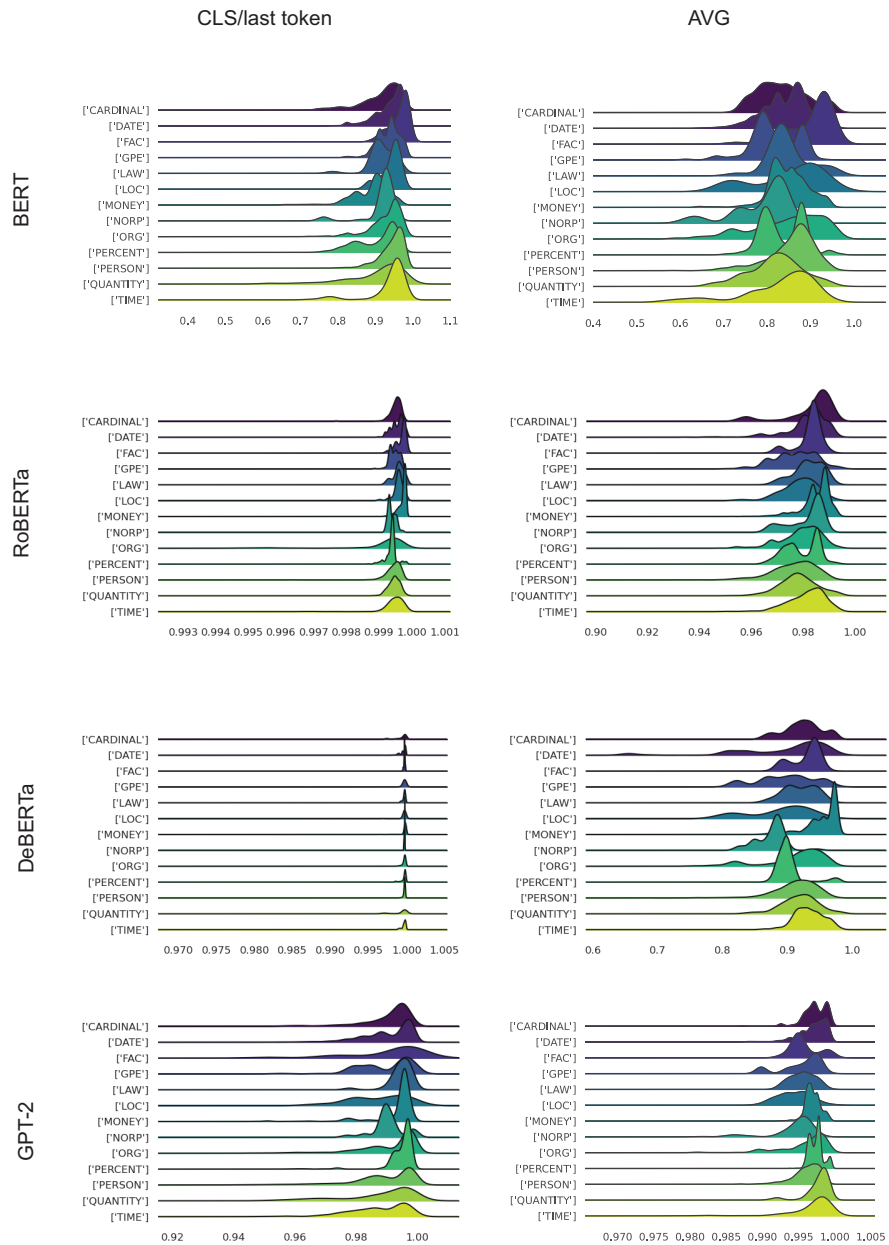}
    \caption{Visualization of distribution of compositionality scores across different types of named entities. }
    \label{fig:named-ents-all}
\end{figure*}

Named entity results can be found in \autoref{fig:ner-all} and \autoref{fig:named-ents-all}.
\FloatBarrier

\FloatBarrier
\clearpage

\section{Frequency and length correlations}
\label{sec:corr-freq-len-human}

\begin{table}[!ht]
    \centering
    \small
    \begin{tabular}{cccc}
        \toprule
        Model and representation & Feature & Spearman $\rho$ & p-val\\
        \toprule
        $\text{BERT}_{\text{CLS}}$ & Word length & 0.2182 & $3.055 \times 10^{-10}$*\\
        $\text{BERT}_{\text{AVG}}$ & & 0.007396 & 0.08722\\
        $\text{RoBERTa}_{\text{CLS}}$ & & 0.01686 & 0.6193 \\
        $\text{RoBERTa}_{\text{AVG}}$ & & 0.3653 & $4.773 \times 10^{-28}$*\\
        $\text{DeBERTa}_{\text{CLS}}$ & & 0.4087 & $1.709 \times 10^{-35}$*\\
        $\text{DeBERTa}_{\text{AVG}}$ & & 0.4484 & $1.340 \times 10^{-42}$*\\
        $\text{GPT-2}_{\text{last}}$ & & 0.3228 & $8.481 \times 10^{-22}$*\\
        $\text{GPT-2}_{\text{AVG}}$ & & 0.0.3125 & $1.719 \times 10^{-20}$*\\
        \midrule
        Human & Word length & 0.05666 & 0.1894\\
        \midrule
        $\text{BERT}_{\text{CLS}}$ & Frequency & 0.2182 & 0.08193 \\
        $\text{BERT}_{\text{AVG}}$ & & -0.08582 & 0.07899\\
        $\text{RoBERTa}_{\text{CLS}}$ & & 0.02548 & 0.9053\\
        $\text{RoBERTa}_{\text{AVG}}$ & & -0.08354 & 0.08193\\
        $\text{DeBERTa}_{\text{CLS}}$ & & -0.1265 & 0.001459* \\
        $\text{DeBERTa}_{\text{AVG}}$ & & -0.2185 & $6.455 \times 10^{-10}$*\\
        $\text{GPT-2}_{\text{last}}$ & & -0.05750 & 0.3595\\
        $\text{GPT-2}_{\text{AVG}}$ & & 0.04382 & 0.5891\\
        \midrule
        Human & Frequency & 0.008363 & 0.9053\\
        \bottomrule
        
    \end{tabular}
    \caption{Correlations of frequency and length with human and model compositionality scores. Corrected with Holm-Bonferroni correction.}
    \label{tab:corr-freq-len-human}
\end{table}

\end{document}